\documentclass{article}

\usepackage{PRIMEarxiv}

\usepackage[utf8]{inputenc} 
\usepackage[T1]{fontenc}    
\usepackage{hyperref}       
\usepackage{url}            
\usepackage{booktabs}       
\usepackage{amsfonts}       
\usepackage{nicefrac}       
\usepackage{microtype}      
\usepackage{lipsum}
\usepackage{fancyhdr}       
\usepackage{graphicx}       
\graphicspath{{images/}}     

\usepackage{times}
\usepackage{epsfig}
\usepackage{amsmath}
\usepackage{amssymb}
\usepackage{multicol}
\usepackage{subcaption}
\usepackage{inputenc}
\usepackage{float}
\usepackage{multicol}
\usepackage{dblfloatfix}

\title{Physics Driven Image Simulation from Commercial Satellite Imagery}
\author{Scott Sorensen$^1$,
        Wayne Treible$^1$,
        Robert Wagner$^1$,
        Andrew D. Gilliam$^1$,
        Todd Rovito$^2$,
        Joseph L. Mundy$^1$ \\
        Vision Systems Inc.$^1$ Air Force Research Lab$^2$ \\
        \texttt{\{scott.sorensen, wayne.treible, robert.wagner,} \\ 
        \texttt{drew.gilliam, joe.mundy\}@visionsystemsinc.com}, \\
        \texttt{todd.rovito@afresearchlab.com} \\
    }

\begin{document}
\maketitle

\vspace*{-7.5mm}

\begin{figure*}[!ht]
\centering
    \subcaptionbox{WorldView 3 image\label{adjusted_real}}
    {\includegraphics[width=0.45\textwidth]{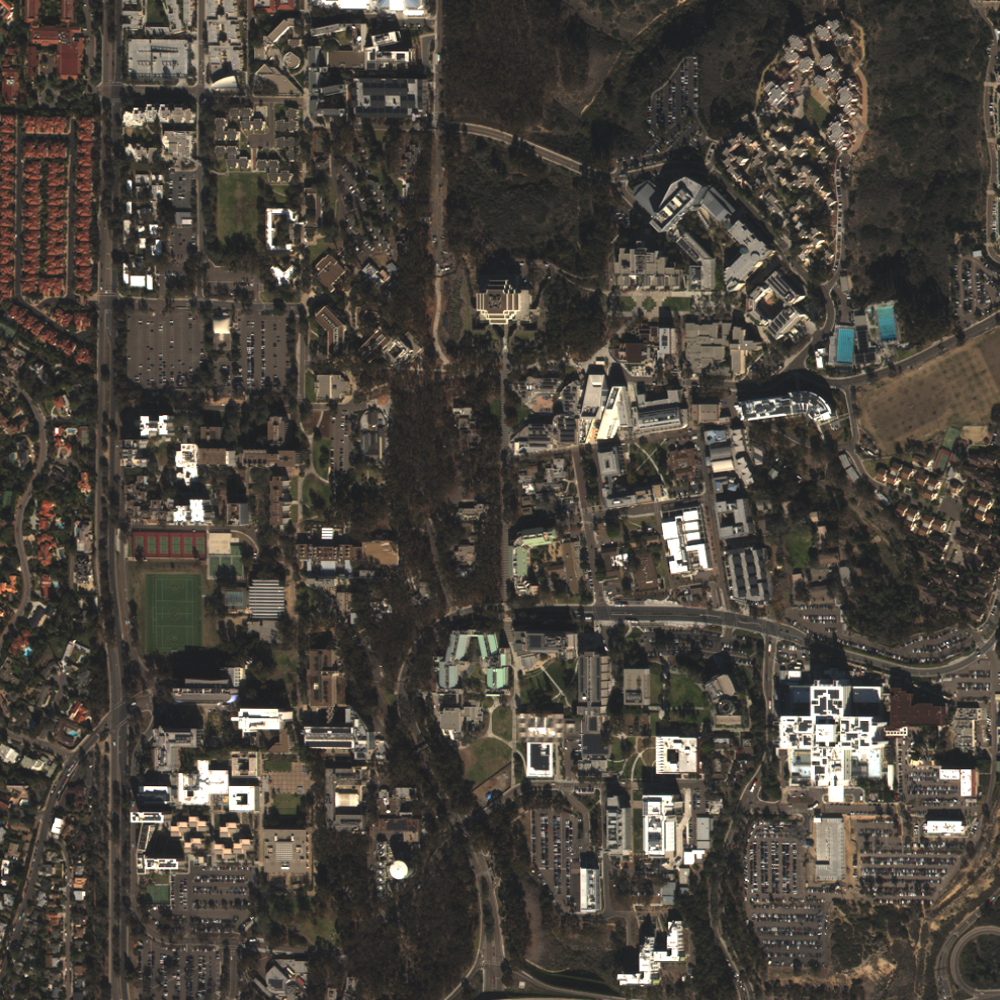}}
    \subcaptionbox{Simulated image}
    {\includegraphics[width=0.45\textwidth]{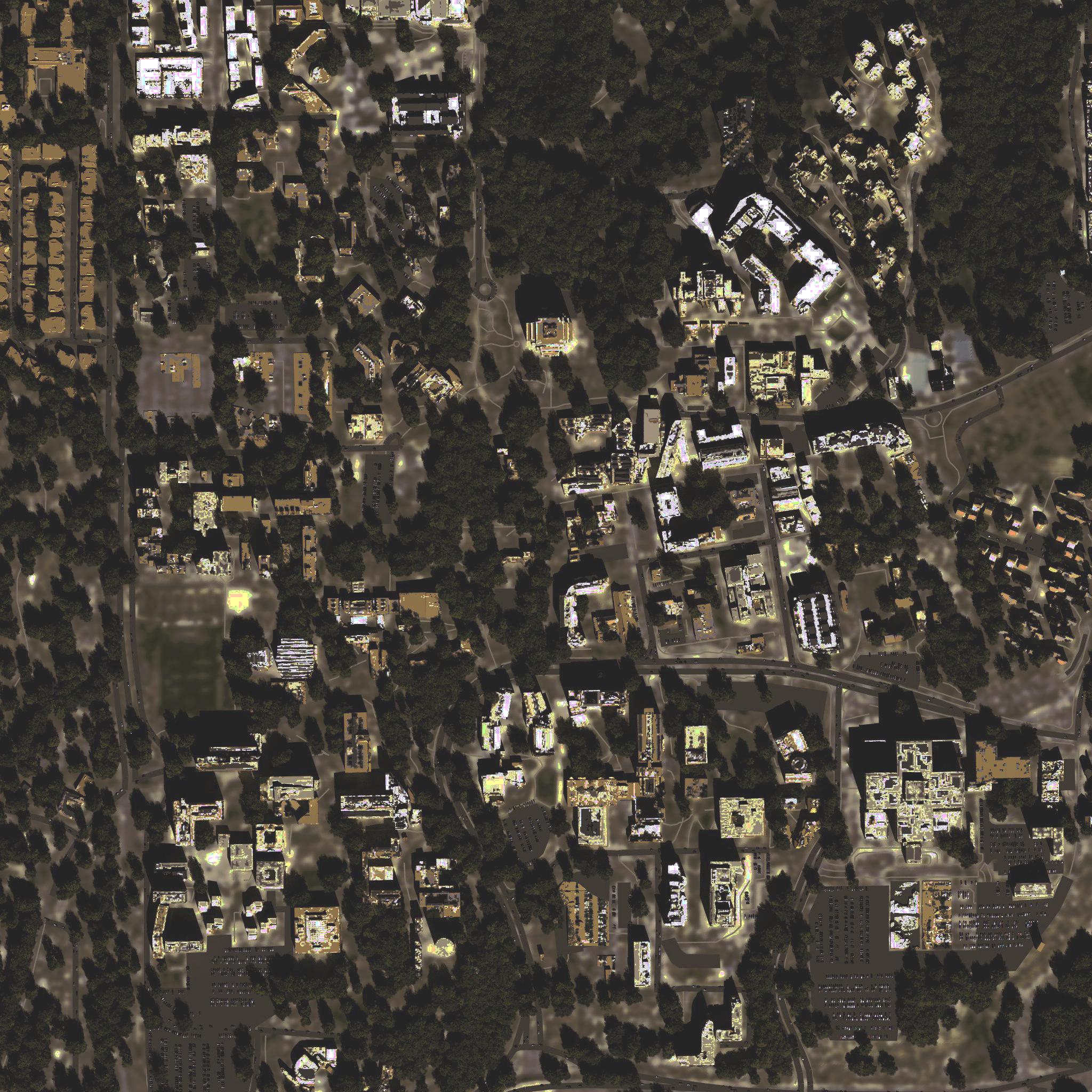}}
    \caption{Simulated and real RGB imagery of a region of University of California, San Diego, showing the simulation closely matches a real WorldView-3 image.}
    \label{fig:sim_and_real}
\end{figure*}


\begin{abstract}
Physics driven image simulation allows for the modeling and creation of realistic imagery beyond what is afforded by typical rendering pipelines. We aim to automatically generate a physically realistic scene for simulation of a given region using satellite imagery to model the scene geometry, drive material estimates, and populate the scene with dynamic elements. We present automated techniques to utilize satellite imagery throughout the simulated scene to expedite scene construction and decrease manual overhead. Our technique does not use lidar, enabling simulations that could not be constructed previously. To develop a 3D scene, we model the various components of the real location, addressing the terrain, modelling man-made structures, and populating the scene with smaller elements such as vegetation and vehicles. To create the scene we begin with a Digital Surface Model, which serves as the basis for scene geometry, and allows us to reason about the real location in a common 3D frame of reference. These simulated scenes can provide increased fidelity with less manual intervention for novel locations on earth, and can facilitate algorithm development, and processing pipelines for imagery ranging from UV to LWIR $(200nm-20\mu m)$.
\end{abstract}

\section{Introduction}
Physics-based simulations using the Digital Imaging and Remote Sensing Image Generation (DIRSIG) framework \cite{8100541} have been widely used to generate synthetic satellite, aerial and terrestrial images for a wide range of applications. DIRSIG's usage is common in prototyping new sensor platforms, generating training imagery, and developing image processing pipelines. Simulation provides data for assessing the performance of image analytics, and is a critical element of many digital engineering pipelines which can advance innovation and modernization\cite{marisa_2021}. DIRSIG can simulate imagery and simultaneously generate per-pixel ground truth labels, enabling machine learning application and the development of a wide variety of other algorithms. Being physics-based and tailored to the remote sensing community, DIRSIG models phenomena outside the scope of many typical rendering approaches, including atypical sensor configurations and frequencies outside of the visible band. These simulations require a level of fidelity beyond other applications to correctly model the complex interaction of materials and radiance across a large range of the electromagnetic spectrum. 

To create a realistic simulation for a given region of earth we will focus on the creation of a detailed scene with appropriate materials. We do not use lidar for geometric reconstruction, so the process begins with the acquisition of satellite images for the region and the creation of a Digital Surface Model (DSM). DSMs are geospatial images that express the elevation of a region, notably including vegetation and man-made structures. Using a DSM we can associate points on the surface with their 3D projection into individual satellite images. This allows us to estimate materials for each image and fuse them in 3D. We assign materials (e.g. concrete, sand, grass, etc) with spectra defined over a broad frequency range to support simulation in a variety of wavelengths. Additionally, we use our material estimates to drive the procedural population of scene elements like vegetation and vehicles. We have built a material library for simulations that includes the ECOsystem Spaceborne Thermal Radiometer Experiment on Space Station (ECOSTRESS) library \cite{ecostress}\cite{aster}, materials re-purposed from other DIRSIG examples, and reconstructed spectra from publications where source data are unavailable.

The DSM is additionally used to compactly model the man-made structures in the scene. We use the reconstructed terrain, modelled man-made structures and populated elements to create a full DIRSIG scene with associated material library and geospatial information. This scene can then be simulated with a variety of sensor platforms. The material definitions are sufficient to simulate the scene in a variety of spectral domains. The geometry of the terrain is constrained to the same resolution as the initial satellite imagery, but the elements which populate the scene can be of arbitrarily high level of detail. This allows for subsequent simulations to realistically capture greater detail than the original images could capture. For example, we can simulate aerial imagery from an off-nadir perspective with a much finer Ground Sample Distance (GSD). We are limited by the source imagery in some ways, but have endeavored to stretch these limitations in resolution and viewing angle. Reconstruction assumes a 2.5D world, meaning overhangs or vertical surfaces are not completely resolved, however scenes on inclines and populated elements can be simulated from many novel viewing angles. We have additionally made use of a mixture model and vector elements to allow for higher resolution material attribution than source images allow for.

Scenes can be simulated in other spectra, for example Long Wave Infrared (LWIR), and DIRSIG can realistically model how the scene has warmed up from the sun over the course of the day \cite{gartley2007polarimetric}. These scenes can be used in a variety of tasks for which DIRSIG simulations are used. For example, a large scene can be constructed to simulate and test a new sampling algorithm or a new imaging platform being developed, or objects can be added to the scene to create training data for a neural network. 

In the remainder of this paper we will cover necessary background information in section \ref{background}. In section \ref{reconstruction}, we will discuss our approach to reconstructing the scene geometry and estimating materials. In section \ref{sim_construct}, we outline how these reconstructions are used to create a full DIRSIG scene suitable for simulation. We will show some sample simulations and discuss the results in section \ref{sim_results}. We will conclude with a some observations in section \ref{conclusion}.
\section{Background}
\label{background}
Physics-based image simulation is used in a variety of applications where more common renderings are insufficient. Physics-based image simulators have been developed for real-time, hardware in the loop, and training applications \cite{mak}, as well as assessment of space object health and status \cite{IDASS}. DIRSIG is a mature simulation program developed primarily for remote sensing applications and has been evolving in capabilities for more than 30 years \cite{8100541}. It is used to simulate single band, RGB, thermal\cite{gartley2007polarimetric}, hyper-spectral\cite{dirsighyper}, lidar\cite{dirsiglidar} and RADAR\cite{dirsigradar}. The system was developed to to assist in hardware designs and to help develop the ground processing pipeline for remote sensing systems. It has been tailored to remote sensing applications with a physics-based approach to handle phenomenology outside of the visible band. 

DIRSIG simulations consist of several parts. The \textit{scene} contains all the solid objects and geometry that interact with light in the simulation. The \textit{atmospheric conditions} simulate how light interacts with the medium of the atmosphere (typically imported from a third party application MODerate resolution atmospheric TRANsmission (MODTRAN)) or using basic models built into DIRSIG. The \textit{sensor platform} consists of a description of the sensor geometry, optics, and spectral response. The \textit{platform motion} describes how the sensor is positioned and moves relative to the scene. The \textit{data collection} describes the specifics of capturing imagery in the simulated scene, such as the time of day and type of task. These elements all go in to a simulation which is driven by a physics-based model of how light from the sun and other sources interact with the materials across the spectrum. In this work, we will focus on the construction of a DIRSIG \textit{scene}. Other components such as detailed sensor descriptions are often proprietary, and are beyond the scope of this work. Simple off-the-shelf models are used in this work and these components are addressed in simulation created in section \ref{sim_construct}.

The input we use to generate a DIRSIG scene is a set of commercially available panchromatic and multispectral satellite images. For locations with available map data (for example crowd sourced data from \cite{wiki:xxx}), we make use of these data sources to improve our simulation. For a given geospatial region we first acquire the images through commercial portals and begin by constructing a Digital Surface Model. Points in the DSM are projected into the original images to evaluate the materials present, and generate a material map. The DSM is also used to compactly model the geometry of man-made structures in the area. We use the material map as well as other geospatial information to logically drive the placement of instances of other objects to the scene. These include trees, other vegetation, as well as vehicles. These elements can be of a much higher spatial resolution, and can have varied materials at an instance level. For example, a scene can be populated with a multiple versions of a single type of car, each instance of that model might have a different color paint, with a corresponding material definition beyond just the visible band.

Other works have aimed to automate DIRSIG scene generation. In his thesis, Lach\cite{LachThesis} extensively researched automating aspects of DIRSIG simulation. This work discussed many different techniques and alternatives for automating aspects of scene reconstruction. Much of the existing work uses lidar for geometric reconstruction and high resolution hyperspectral aerial imagery for material assignment. Similarly \cite{6528215} aligned hyperspectral imagery with lidar to create a simulated natural scene. Unlike these works, our approach uses no aerial data of any sort, and no lidar for high accuracy geometric reconstruction. In \cite{LachThesis} the author states "the unavailability of lidar data poses perhaps the greatest hindrance on performing semi-autonomous scene construction." We overcome these issues by creating and leveraging a high resolution and accurate DSM constructed from satellite imagery with global availability, as well as an in depth reconstruction method that compactly models buildings, and handles underlying terrain. Costly lidar is not available for large areas of the earth, and the proposed approach still allows for detailed simulations to be constructed for these areas.

\section{Reconstruction}
\label{reconstruction}
In this section we will discuss the process of extracting information from the satellite imagery and other data sources. The sources used in this work are shown in table \ref{table:data_table}. We use commercially available Maxar images. From these images we extract scene geometry and physical makeup of the materials in the scene. This process begins with the creation of a DSM. This DSM is used for a 3D frame of reference throughout the rest of reconstruction. It is used in material estimation to allow for accurate projection of a 3D scene into each image. It is also required for modelling man-made structures and to create a Digital Terrain Model (DTM) which is used for the bare earth surface in our simulated scene.

\begin{table}[ht]
\centering
\resizebox{\textwidth}{!}{%
\begin{tabular}{|c|c|c|c|c|}
\hline
Data source & Type & Quantity \\ 
\hline
WorldView-3 imagery & Raster Images &  88 Panchromatic \& Multispectral images \\
OpenStreetMap$^*$ \cite{wiki:xxx} & Crowd sourced vector data and labels & Query results from geographic region \\
ECOSTRESS\cite{ecostress}\cite{aster} & Material library with spectral reflectance curves & Full library \\
Materials sourced from literature$^*$\cite{solar}\cite{terracotta} & reflectance curves published in other works & 2 material reflectance curves\\ 
\hline
\end{tabular}%
}
\caption{A table outlining the various sources of data to create the simulation in this work. $^*$\scriptsize Indicates that this data source is optional }
\label{table:data_table}
\end{table}

\subsection{Digital Surface Model Creation}
Our work does not rely on lidar, so a DSM forms the basis of the 3D geometry of a reconstructed scene. Creation of a DSM is an involved process and beyond the scope of this work, but we will outline the process here and direct the reader to \cite{abs-2104-04843} for a recent evaluation of the process. To create a DSM the scene is broken into tiles that ensure rays from the camera are parallel and the images are geocorrected. This process aligns imagery which may have registration errors on the scale of a few to tens of meters. We use an automatic approach outlined in \cite{journals/ijcv/OzcanliDMWHT16}. This method simultaneously registers multiple images allowing for accurate 3D reconstruction. A subset of images is selected based on viewing angle, ensuring there is some difference in view between each image, however it is not so great that stereo matching is likely to fail. Individual pairs of images are rectified and stereo matching and reconstruction techniques are used. We use semi-global matching \cite{conf/cvpr/Hirschmuller05} for disparity estimation, which is widely used in the geospatial domain \cite{https://doi.org/10.1029/2018EA000409}. A detailed analysis of this approach can be found in \cite{conf/cvpr/OzcanliDMWHT15}. These pairwise models are fused using a probabilistic approach building off the works of \cite{8354141}. This process yields accurate DSM at a resolution of $0.3$m. 

\subsection{Material Estimation}
We conduct material estimation in two stages in our overall pipeline with early coarse classification followed by a later step of fine-grained classification. The initial stage is used for broad category classification that allows for more accurate geometric reconstruction. The second stage of material classification yields instance level classifications with a much larger corpus of reference spectra.

For coarse classification, multi-spectral imagery taken over the region defined by the DSM can be matched with reference spectra of known materials to produce a pixel-wise material map. The material map is useful for identifying regions of vegetation, rocks, soil, water bodies, and man-made structures. Areas with water or trees are dynamic, low texture, and are difficult to match during stereo reconstruction. Consequently these areas can be quite noisy in the DSM. By classifying forested regions and water bodies we can apply simple filtering to the DSM prior to 3D structural modeling, as those regions can confound the 3D modeling process. These regions are re-populated at later stages in the simulation process.

The reference spectra used for this stage includes only 10 categories of material ranging from common types such as vegetation, water and asphalt. These labels follow from the IARPA CORE3D challenge, and the spectra originate from the ASTER spectral library, \cite{aster} which is part of the larger ECOSTRESS library. These were hand picked semantically meaningful labels that represent a very large portion of the earth's surface, especially the portions we are interested in modelling, namely regions with man-made structures as well as natural terrain. The coarse classification is useful in geometric modelling, but a more fine-grained material map is needed to make a realistic simulation, and this requires accurate estimates for materials from a site-specific sample collection or a large material library.

Fine grain material estimation is a more involved process and requires radiometric calibration, reference material matching and vector quantization. These techniques are outlined in  \ref{radiometric_calibration}, \ref{reference_matching},  and \ref{vector_quantization} respectively.

\subsubsection{Radiometric Calibration}
\label{radiometric_calibration}
Radiometric calibration is the process of converting raw sensor data to meaningful physical quantities such as radiance and surface reflectance. The intent of this process is to produce a consistent image/target appearance across a variety of platforms and atmospheric effects suitable for comparison to well-known material spectra. Calibrated imagery is typically produced through a combination of vendor specified calibration parameters and scene-specific calibration routines.
 
Unfortunately, standard calibration techniques often still produce significant variation in surface reflectance estimates that result in poor material classification. This is alleviated by refining the calibration step using pixels of known material (e.g., road asphalt). Road pixels are identified using a digital road map such as OpenStreetMap (OSM), and an additional calibration adjustment is computed which minimizes the difference between the reference material spectra and the multi-spectral image samples.

\subsubsection{Reference material matching}
\label{reference_matching}
A common method for comparing multi-spectral image samples to surface material reference spectra is known as the Spectral Angle Mapper (SAM)\cite{KRUSE1993145}. The goal of this method is to find the closest match between the image pixel, $\vec{x}$, and a surface material spectra, $\vec{r}$, by minimizing the distance between the two vectors using the cosine distance function given in Equation \ref{eq:sim}.
\begin{equation}\label{eq:sim}
\alpha = \cos^{-1} \left( \frac{\vec{x}\cdot \vec{r}}{\lVert \vec{x} \rVert \cdot \lVert \vec{r} \rVert} \right)
\end{equation}
The material match with smallest angle $\alpha$ is taken as the material type for a specific pixel.

This mapping can be applied to every pixel in the image to produce a new image that represents a material map. This method is most successful when the scene materials are known a priori using resources such as a digital road map, a geological map, and/or vegetation map that provide detailed knowledge of what materials are present in the region of interest so that a site-specific reference material library can be established.  A focused reference library will help reduce unrealistic false positives such as labeling pixels of a material map located in the desert as ice.

If such resources are not available for an area of study, it is possible to use a large corpus of reference spectra to perform the material mapping. However, slight variations in the observed satellite data may lead to similar samples being misclassified as belonging to different material classes. For example, neighboring pixels in a desert region may be classified as limestone or sandstone, when only sandstone exists at that location.  To reduce misclassification, we quantize the multi-spectral image so that only a small number of pixels need to be matched to the reference materials.

\subsubsection{Multi-spectral image color quantization}
\label{vector_quantization}
Color quantization is a well known method for reducing the number of colors required to show an image, while still preserving the overall appearance quality.  For instance, it can be used to convert a 24 bit RGB color image to a 8 bit color image.  This is accomplished by performing a pixel-wise vector quantization of the image, where pixels are represented $\mathbb{R}^3$ and K-means clustering is used to find 256 color clusters.  The cluster centers define the new color palette for the 8 bit image.  

Vector quantization can be applied to multi-spectral image data to reduce the variation in the observed data.  Instead of matching each pixel to a specific material in the reference library to construct a material map, we can perform a pixel-wise vector quantization of the multi-spectral image to create a set of clusters that have similar reflectance properties that can be mapped to the same material.

Formally, WorldView-3 multi-spectral imagery contains eight bands of digital reflectance data ranging from 400-1040 nm.  Specifically, the spectral ranges are Band 1: Coastal (400 - 450 nm), Band 2: Blue (450 - 510 nm), Band 3: Green (510 - 580 nm), Band 4: Yellow (585 - 625 nm), Band 5: Red (630 – 690 nm), Band 6: Red Edge (705 - 745 nm), Band 7: NIR1 (770 - 895 nm), and Band 8: NIR2 (860 - 1040 nm).  This type of multi-spectral image is typically referred to as a VNIR (Visual + Near Infrared spectrum) image. 

Each pixel of a calibrated VNIR image can be cast as feature vector, $x_i$, as
\begin{equation}\label{eq:vnir_pixel}
x_i = [b_1, b_2, b_3, b_4, b_5, b_6, b_7, b_8],
\end{equation}
where $b1 \cdots b8$ are the band intensity values at the pixel location.  The feature vectors are clustered in $\mathbb{R}^8$, using k-means++ clustering \cite{arthur1283494}.   In practice, we subsample the original VNIR image prior to clustering to further reduce the computational cost.

A single channel image of the same size as the VNIR image is formed by mapping the cluster index associated with the image pixel to the position in the original VNIR image.  We refer this this as the class map.

The class map can be converted to a material map by building a catalog of materials using the cluster centers ${\mathcal{C} = {\{c_1, c_2, \cdots, c_k\}}}$.  Specifically, each $c_j$, defined by  
\begin{equation}\label{eq:cluseter_center}
c_j = [b_1, b_2, b_3, b_4, b_5, b_6, b_7, b_8]
\end{equation}
is matched with the most likely material candidate found in the reference library using SAM to create the material catalog ${\mathcal{M} = {\{m_1, m_2, \cdots, m_k\}}}$.  Note that multiple cluster centers can be matched to same material based on the quality of the spectral data and the number of reference samples, resulting in a material catalog that is smaller than the number of clusters.  Therefore, the material catalog is remapped to use only the unique material set ${\mathcal{M'} = {\{m_1', m_2', \cdots, m_n'\}}}$, where $n \le k$.

\begin{figure}
\centering
    \subcaptionbox{Quantized multi-spectral image patch $(k=50)$ \label{ucsd_clustered_patch}}
    {\includegraphics[width=0.23\textwidth]{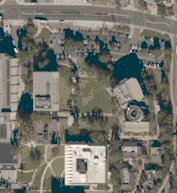}}
    \subcaptionbox{Matching ECOSTRESS image patch (material count = 38) \label{ucsd_ecostress_patch}}
    {\includegraphics[width=0.23\textwidth]{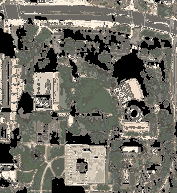}}
    \caption{Fine-grained material matching for small region of the UCSD campus.}
    \label{fig:fine_grained_material}
\end{figure}

For fine-grained material matching we used the ECOSTRESS library \cite{ecostress}, which consists of more than 3400 different spectra of common surface materials. An example result from the UCSD campus is shown in Figure \ref{fig:fine_grained_material}, where the quantized multi-spectral image is shown as a color image on the left and the corresponding region using colors derived from the matching ECOSTRESS materials on the right. For this image, we used $k=50$, but there are only 38 unique materials (i.e. $n=38$).

The material maps created with this method are used to create DIRSIG material maps, DIRSIG material mixture maps, and drive procedural object placement.

\subsection{Structure Modelling}
Our approach to modelling man-made structures aims to accurately capture the geometry while reducing the memory footprint. This portion of the pipeline is also outside the scope of this work, however this component of our system addresses the same problem as the IARPA CORE3D challenge, and there are a variety of works addressing this question \cite{8899241}\cite{https://doi.org/10.48550/arxiv.2005.09223} including the open source work of \cite{conf/cvpr/LeottaLJZLS0LZC19}. We will briefly summarize our approach here. The DSM is created from satellite images and formed as a 2D raster, this means only the top surface of any structure is captured. This is sufficient for most the vast majority of structures that have walls and do not feature undercuts. We aim to compactly represetn thses structures in terms of primitives with few vertices, mostly straight edges, and mostly $90^\circ$ and $45^\circ$ angles. In practice this means extruded polygonal regions, representing different roof planes. Since most man-made structures feature straight lines and right angles we favor these to mitigate error propagation from noise in the DSM. The resulting extruded polygons should therefore consist of few mostly orthogonal edges, and closely match the real world structure.

The process of modelling begins with the DSM, which can be represented as a point cloud in a Local Vertical Coordinate System (LVCS), which is a Euclidean space with a local origin. We can easily translate 3D points from a global (latitude, longitude, elevation) coordinate frame of reference to the local (x,y,z) coordinate frame. We convert every pixel in the DSM to a 3D point in our LVCS. We utilize a Random Sample And Consensus (RANSAC) based primitive fitting method to fit geometric primitives to identify candidate shapes \cite{schnabel-2007-efficient}. This technique allows us to specify priors for shape fitting, such as shape parameters, and priors about fitting. We have experimented with a variety of primitives but have found planes fit man-made structures best with fewer false positive candidate shapes. We have specified priors about the orientation of the plane (planes within a range of reasonable roof slopes ($<50^\circ$)), priors about size (planes of appropriate roof sizes (at least 75 points)), as well as priors on noise and quality ($\epsilon = 0.35$, bitmap$\epsilon = 0.35$ (priors on plane fitting and shape fitting)). Each candidate plane consists of the parameters defining the plane and the set of points which have met our fitting criteria.

We then begin the process of creating planar regions, polygons within the candidate planes that fit the pointset. We begin by projecting points into the plane and converting to a 2D planar coordinate system using basis angles. We then compute an alpha boundary of the projected 2D pointset \cite{akkiraju:alpha}. An alpha boundary is a generalization of a convex hull that allows for non-convex regions. We favor using a 5m $\alpha$ parameter which prioritizes longer edges. The resulting boundary can still be noisy and is prone to self intersections, 1D edge chains and other issues. We correct these by splitting self intersecting boundaries into multiple regions and eliminating regions that do not meet size criteria. 

We compute an edge map using \cite{10.1109/TPAMI.1986.4767851} for the most nadir satellite images and project these edges into each fitted plane to further refine the polygonal boundaries. By examining the predominant orientations of projected edges we can determine the orthographic axes for a given region. These edges are used to snap boundary points to edges that better reflect the predominant roof edges in man-made structures. We then take these refined planar regions and extrude them to the ground, creating triangular meshes. We then compute UV coordinates (coordinates that map 3D vertices to a 2D image for texturing using basis vectors U and V) to allow us to apply material maps in simulation.

\subsection{Digital Terrain Model}
To create a good simulation we treat the terrain separately from other elements. To this end we create a DTM, which is similar to a DSM except it does not model the elevation of vegetation or man-made structures, but just the underlying terrain. To create a DTM we start with the filtered DSM point cloud (without trees and water). The tiles used in DSM construction are used to separate regions of the scene, and we compute a single initial elevation value for each. This is done by computing a histogram of elevations for the tile and determining the most probable ground elevation. This is computed by identifying the minimum histogram bin with more than 1\% of elevation values. 

We then compare adjacent tiles to create a smooth surface and eliminate values that represent the elevation of buildings or other non-terrain elements. Minimum elevation values of adjacent tiles are averaged to create elevations at corners of the neighboring tiles. For each tile 4 planes are computed by fitting a plane to 3 of the 4 corners. We then determine how closely the remaining point fits the plane and eliminate outliers that do not fit the tile's planar approximation, and replace it with the elevation of the plane at that point. This process is continued across each tile to refine estimates of the corners. The process is carried out iteratively, refining each corner between tiles. In our work we found 10 iterations resulted in a smooth surface that does not represent the elevation of trees or buildings.

The corner points of this smoothed terrain approximation are then connected into triangles to form a mesh surface. To allow for material mapping we compute UV coordinates for each vertex, associating the 3D position with 2D image coordinates in the material map. 

\section{Simulation Construction}
\label{sim_construct}
With the creation of detailed material maps, extruded planar buildings and a DTM mesh, we can begin assembling a simulation. This process involves populating the simulation with instanced scene elements based on material designation or other geospatial data sources. We have experimented with a few techniques for applying materials to the scene, from direct translation, to material remapping, to creation of material mixture maps. These techniques are combined to create a full DIRSIG scene.

\subsection{Procedural Element Population}
A detailed environment consists of much more than terrain and man-made buildings and large scale structures. Trees, undergrowth, cars and trucks are just some of the elements we can commonly see in satellite and aerial imagery. To create realistic simulations we need to include elements like these in abundance. DIRSIG supports instanced scene geometry for this purpose. Instancing allows for a great number of variants of a base object to be used with little overhead. The base object can be copied with different translation, rotation, scaling, and even material attributes. DIRSIG automatically handles placing the objects on the terrain and handling elevation changes throughout the scene. This allows us to densely populate a forested area with just a few unique models of different trees, but placing many copies at different orientations, and with different scaling. In the case of vehicles, we can populate roads and parking lots with the same car in different paint materials. With just a few models we can create a large number of diverse vehicle instances. 

To create instanced geometry for the scene we have used two different approaches. We have specified the explicit placement parameters, and created a density map that specifies regions to populate at various densities. These different approaches share some common features but are better suited to specific use cases. Density maps are used in DIRSIG to allow it to populate elements at different densities driven by an image. By creating an image and using the UV coordinates, we can add elements in greater and lesser abundance to different parts of the scene. This is especially useful for adding geometry to the scene where we have simple material labels. For example, we would not be able to reconstruct small shrubs and other vegetation but can use material classification results to specify that vegetation should be placed on soil areas in higher abundance. By creating a density map with no or low vegetation in regions with rocky terrain and a high abundance of vegetation in other regions we can effectively populate the area with small scale vegetation.

By specifying placement parameters, instanced objects are placed at a particular points in the scene with a specific orientation etc. This is useful for placing objects according to some more complex rule set. For example cars are placed at the correct orientation and position on roads in our scene using road data from OSM\cite{wiki:xxx}. We still populate these positions with randomized models from a specified list. 

To correctly place the cars, we take each road, extract the edges, and translate them to Euclidean coordinates in our LVCS. Given an edge representing a section of road we compute the orientation vector $v$ by normalizing the vector between the vertices $v_0$,$v_1$ defining the edge. We then compute an angle $r$ clockwise and orthogonal to this vector using equations \ref{eq:right_angle} and \ref{eq:right_vec}. This yields a vector pointing to the right of the direction of the road section. Using a lane spacing of approximately 4 meters for the USA, we can place vehicles on the right side of the road and in the correct orientation. We negate the vector and reverse the orientation to put cars in both directions along the road at minimum intervals with a low probability of any point on the road being occupied.

\begin{equation}\label{eq:right_angle}
    \theta=atan2(v_0,v_1)-\frac{\pi}{2}\\
\end{equation}

\begin{equation}\label{eq:right_vec}
     r = [\cos(\theta),\sin(\theta)]
\end{equation}

\subsection{Decal Map Generation}
In addition to populated elements we can incorporate vector information into high resolution simulations in the form of geometry applied as decals. The road network as detailed in OSM map data is a network of nodes and edges that can be easily converted into planar segments. DIRSIG allows this geometry to be applied as a decal with its own materials. By generating planar geometry with asphalt material we can include roads as geometric elements at high resolution without aliasing. We create simple meshes for each segment of road with uniform lane width. These simple polygonal regions are assigned asphalt material for the road network and footpaths can be added in a similar fashion with concrete materials.

\subsection{Material Reconstruction, Translation, and Remapping}
After material classification we can begin to apply materials to the DIRSIG scene. DIRSIG supports material maps, a form of texture mapping with an image containing the indices of materials in the simulation's material database. In our work this process can take the form of direct translation or remapping to other materials in the database. In the direct translation case we can simply use the reference spectrum for each material class. This process is straightforward, as DIRSIG supports a simple reflectance lighting model which allows us to use the reference spectrum with a simple unit conversion. 

Remapping the materials allows us to use a library of existing materials, and associating those materials with classified labels. This is useful in a case where good reference spectra are not well known, or in cases where we choose to use a defined composite class like sand instead of a mix of all the different component materials that make up sand. In this example we could map a variety of classes into the single label `sand'. We also employ remapping for contextual materials. For example, we may coarsely classify parts of the scene as asphalt, but in reality, there are different asphalt types for roads and shingle roofs. We can remap each of these labels separately depending on whether the surface is part of the terrain or man-made structures. Additionally remapping may be useful for materials that match incorrectly. For example, in a simulation we constructed, a portion of the ground in a scene in North America matches to the reference spectra of a species of pine that is indigenous to Morocco, but was remapped to `grass'.

In some cases appropriate reference spectra are not available in either existing publicly available simulations or other spectral libraries such as ECOSTRESS \cite{ecostress}\cite{aster}. Spectral reflectance of various materials is important for a variety of scientific and engineering tasks, so many researchers and manufacturers have conducted measurements for various materials. These measurements are often presented graphically in publications and spec sheets for materials, but rarely are the raw measurements available. To fill the gaps in our material library, we have created a utility which allows us to trace the spectral curves in image and PDF sources to reconstruct spectra. A user uses a simple graphical user interface to click the axes and enters the extrema of a graph and then traces the curve with clicks to recreate the source curve. This allows for us to integrate new materials into simulation based on a wide variety of public data. While these reconstructed spectra are not of the same precision as a direct measurement, as they are limited by the source image resolution and by manual exactitude, they allow for quick introduction of new materials.

\subsection{Mixture Map Generation}
While the material maps discussed so far allow us to capture and simulate materials in a scene, they suffer from poor resolution when simulations are of a greater resolution. The labels we assign to a material cannot be of greater resolution than the source imagery, but simulated imagery can have a GSD of completely different scale. As a result there are clear pixelation and aliasing artifacts. In DIRSIG, points on the surface of an object do not need to be labeled with a single material, but can consist of multiple materials of differing abundance based on a mixture map. A mixture map in DIRSIG is a multi channel image with relative abundances of different material in each channel. We have opted to create a mixture map of a higher resolution to not only increase effective resolution but allow for smooth transitions between material classes.

To create the mixture map, we naively up-sample the material map, duplicating pixels with no blurring or other typical re-scaling techniques to ensure class labels are unchanged. We create a single channel image for each material class in the up-scaled map. Each of these single channels has a 1 wherever that material is present in the material map, and a 0 everywhere else. We then apply a Gaussian kernel to the images to blur the boundaries and expand the regions labeled as belonging to that class while decreasing the probability. This leads to multiple overlapping materials around boundaries, but DIRSIG expects the total probabilities in a mixture map to sum to one, so we recompute total probabilities by normalizing every pixel across all the single channel images and combining this into a single array. This is then exported as a multi channel ENVI image for use in simulation.

\subsection{Simulation File Generation}
While perhaps less algorithmically interesting, an important part of our system is the automatic and semi-automatic generation of DIRSIG files. This allows us to directly include files we generate, and facilitate quicker construction of the full simulation. The key DIRSIG files we have generated can be broken up into geometry, materials, and maps in DIRSIG terminology. In terms of geometry, DIRSIG supports the common `.obj' 3D model standard, so this is used for models we create. We further write text files with information on procedural object placement and create and modify `.glist' files to incorporate these models into the scene. Materials spectral curves are written to simple reflectance `.txt' and emissivity `.ems' files and a material database '.mat' file are also created. The maps created can vary depending on the scene, but can contain material and abundance maps in traditional image formats such as `.png' or `.tif' or in the case of material mixture maps in the form of ENVI `.img' files. Tying everything together is the DIRSIG '.scene' file. In practice these are often generated separately and manually tied together, as we have been experimenting with different aspects of the pipeline, however we have created full simulation ready scenes automatically for simpler scenes.

\section{Simulation Results}
\label{sim_results}
\begin{figure}
    \centering
    \includegraphics[width=0.45\textwidth]{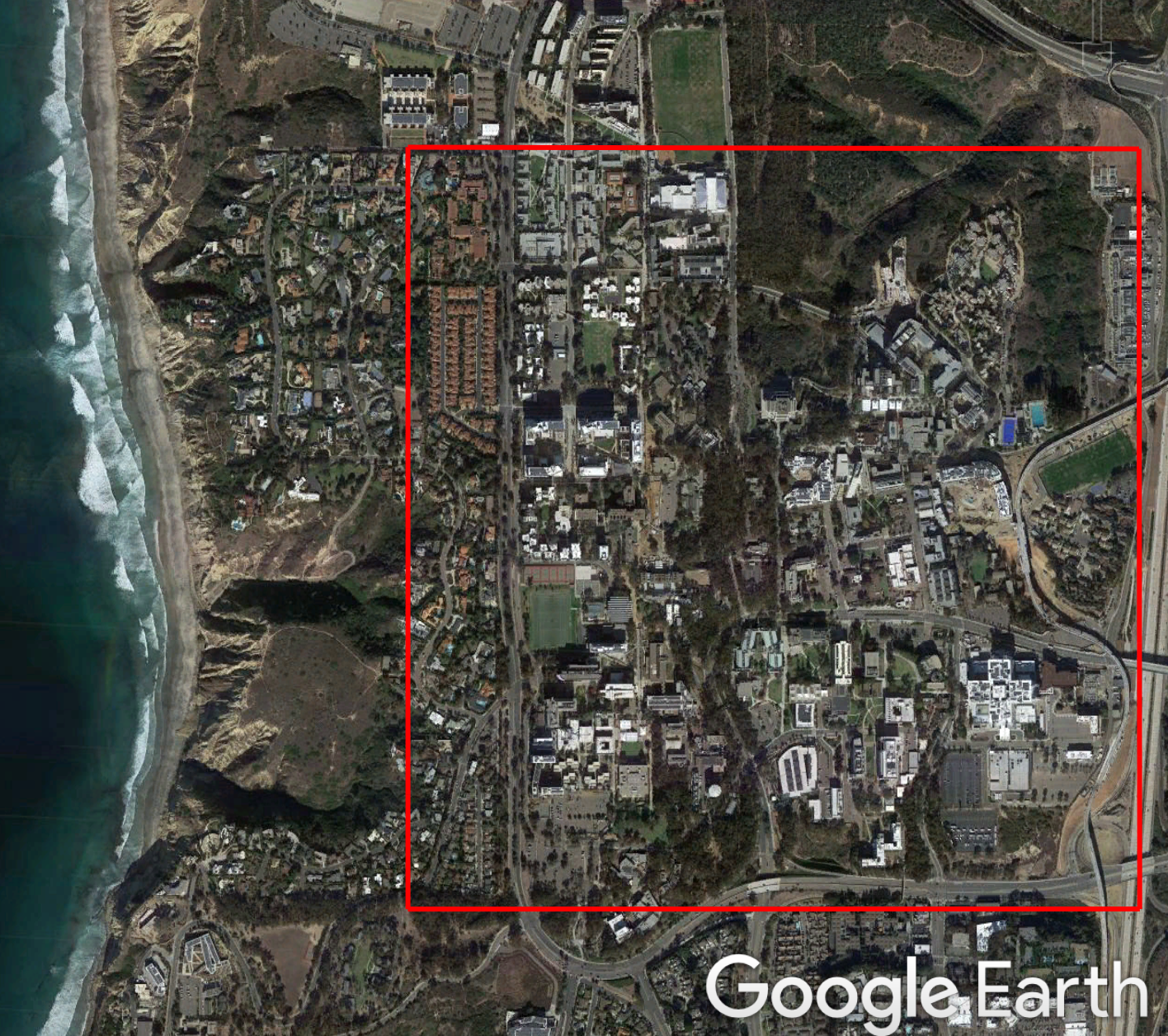}
    \caption{A Google Earth screenshot of the simulation area. Approximately $2.75km^2$ of the UCSD campus in California.}
    \label{fig:google_screenshot}
\end{figure}

To illustrate our technique we have created a simulation of a region of University of California San Diego's campus (UCSD). This region was selected because it contains a mixture of different terrain materials, elevation changes, vegetation, as well as man-made structures including buildings, roads and parking lots. The region is shown in Fig. \ref{fig:google_screenshot}. The location has relatively accurate crowd sourced labels available from open street maps, as well as commercial satellite imagery in the CORE3D public dataset was provided courtesy of DigitalGlobe (now Maxar). This dataset was created for the IARPA CORE3D program\footnote{\url{https://www.iarpa.gov/index.php/research-programs/core3d/}} and distributed by Spacenet \cite{spacenet}. The area we have reconstructed for simulation consists of 25 tiles occupying approximately $2.75km^2$. The region includes the campus center, academic buildings, recreational facilities, ecological park, surrounding housing, roads and parking lots. We have constructed the simulated scene semi-automatically, by iterating on the simulation and addressing issues as they occur. This process is time consuming but represents a drastic reduction in work over a purely manual process.

\begin{figure}
    \centering
    \includegraphics[width=0.45\textwidth]{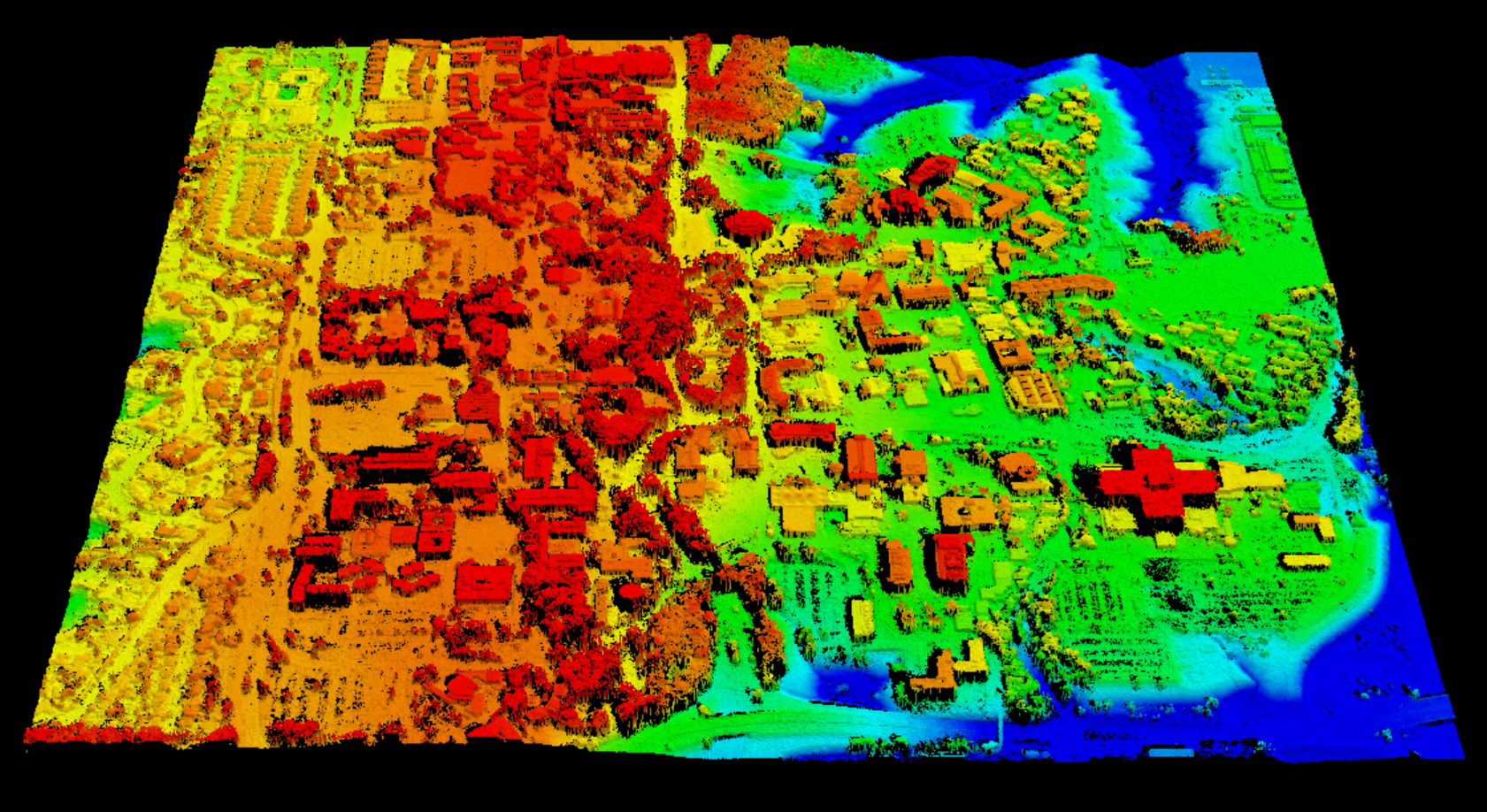}
    \caption{The colormapped and shaded DSM of the simulated region}
    \label{fig:DSM}
\end{figure}

\begin{figure}
    \centering
    \includegraphics[width=0.45\textwidth]{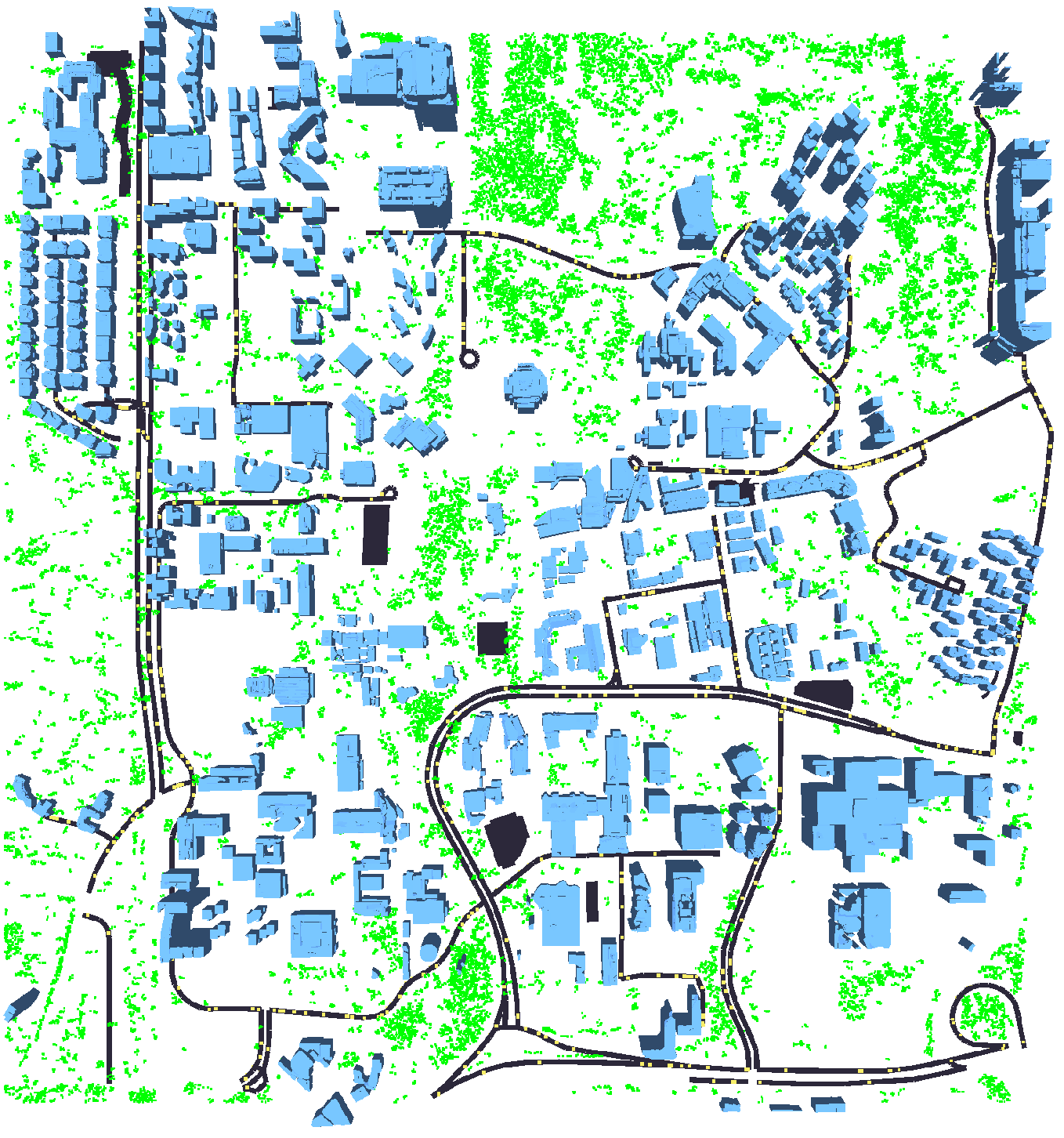}
    \caption{An illustration of some of the components that create the simulation. These include modeled buildings (blue), road network(black), populated trees (green), and the cars populated on the road network (yellow)}
    \label{fig:elements}
\end{figure}

\subsection{Geometry}
 As described we begin by constructing a DSM. Fig. \ref{fig:DSM} shows the $0.3$m resolution DSM created using our process. This DSM was constructed using 41 of the geocorrected images selected due to viewing angle. Structural modelling is performed resulting in compact building models (the blue shapes in Fig. \ref{fig:elements}). This process also creates a DTM mesh which is used as the terrain for simulation. The DTM and modeled structures have UV coordinates computed using the camera parameters associated with the projection of materials maps as computed as described in subsection \ref{sub:sim_materials}.

\subsection{Materials}
\label{sub:sim_materials}
To create the simulation we have used a variety of sources including about 35 materials from the ECOSTRESS library, and about 30 materials from existing example DIRSIG simulations (including all vehicle and tree materials). We have manually reconstructed spectra for monocrystaline solar cell with 3.2mm tempered glass\cite{solar}, and unweathered Milanese terracotta tiles\cite{terracotta}, as there was no suitable replacement material in our material database. For the terrain we have utilized the full mixture map approach with limited remapping as smooth transitions between material types works well on natural terrain such as grass, trees, rocks and sand. We achieved this by doubling the resolution of the initial material map in both dimensions to upscale. We have remapped some exotic terrain materials (for example the aforementioned Moroccan pine) and have used more sand materials. The man-made structures feature lots of small details and greebles (details that breaks up a larger form) of different materials. We found that keeping a full resolution single material map preserved these details better and looked more realistic when simulated at the same approximate resolution.  

\subsection{Populated Elements}
We use the material estimates to drive populated element geometry placement. Many of the materials in the material map were successfully identified as tree canopy materials, many of which were even native to the Southern California region we are attempting to simulate. While it would be ideal to simulate these exact trees and would result in the most accurate simulation, creating a geometry and material library for a wide variety of tree types is beyond the scope of this work. We have used a few types of oak, maple and dogwood models. These models are placed in areas labeled with tree materials and can be seen as the green spots in Fig. \ref{fig:elements}.

We additionally populated the scene with cars. These cars are placed on the road network and in parking spots. The cars are placed on roads facing in the proper direction for right hand drive roads. These cars are shown in yellow in Fig. \ref{fig:elements}. We have also  manually annotated the layouts of about 15 parking lots by drawing lines across the rows in Google Earth. In the scene we populate 70\% of parking spaces with vehicles. The road network (shown in black in Fig. \ref{fig:elements})  and walking path network are converted into decal maps for vector geometry with their associated materials.

\subsection{Simulation}
The various geometric and material elements were combined into a scene suitable for DIRSIG simulation. We have created simulated imagery using a simple RGB and Near-Infrared (NIR) perspective camera with a narrow field of view positioned above the scene. As our goal has been \textit{scene} production we have not committed a large effort to modelling a sophisticated \textit{sensor platform}, accurate \textit{atmospheric conditions}, or other components of a full synthetic imaging pipeline. We have used DIRSIG's tools to create a simple perspective camera, and used a basic mid-latitude summer atmosphere. Sensor response curves, artifacts related to motion, preprocessing, and more sophisticated atmospheric effects are not modeled.

We have not created a sensor with a response that simulates an existing platform, so for illustration purposes we have compared an image captured by Worldview 3 (a) of this region to our synthetic image (b) in Fig. \ref{fig:sim_and_real}. This comparison was made by matching the histogram of the synthetic image to the real image. This helps mitigate any response issues for a more direct comparison, but does mean there are some localized color inaccuracies as the histogram is computed globally. Furthermore, it should be noted that the satellite uses a push broom camera, and we have used a simple perspective camera in our simulation. We argue that the two scenes look quite similar, and demonstrate the capabilities of our approach. There are, however a few notable limitations with our approach.

Our simulation approach has many components each of which can contribute errors. Geometric errors can occur via failures in reconstruction. One cause of this is a dynamic real world scene. The UCSD campus has featured a number of construction and renovation projects over the course of our observation and reconstruction window. This results in areas with a poorly fit DSM that does not reflect the instantaneous snapshot of the elevation and structures of the scene. Our material estimates are limited by the observations and material libraries we have available. Furthermore, we carry out matching based on the entire observed spectrum, which may de-emphasize the visible portion of the spectrum. Pigments, dyes and paints are used to alter the color to human observers, and do not necessarily affect reflectance outside of the visible band. Populated elements rely on labeling, either from material driven estimates or from other sources. In this scene, several parking lots were not labeled or miss-attributed in Open Street Maps.

\begin{figure}
    \centering
    \includegraphics[width=0.45\textwidth]{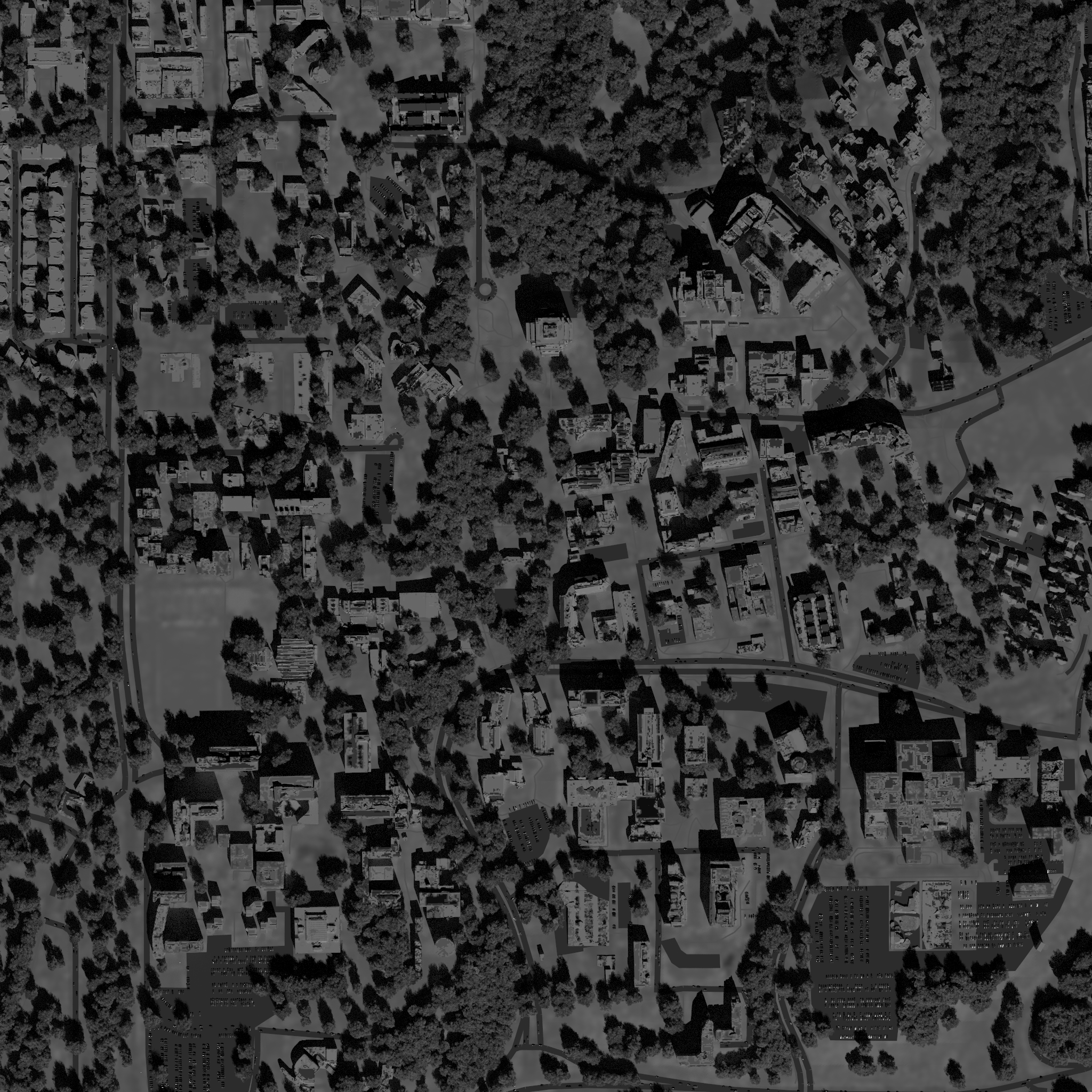}
    \caption{A Near Infrared ($0.7 - 2.5\mu m$) simulation of same region of UCSD}
    \label{fig:NIR_sim}
\end{figure}

\begin{figure}
\centering
    \subcaptionbox{RGB simulation showing fine level of scene detail \label{rgb_lot}}
    {\includegraphics[width=0.23\textwidth]{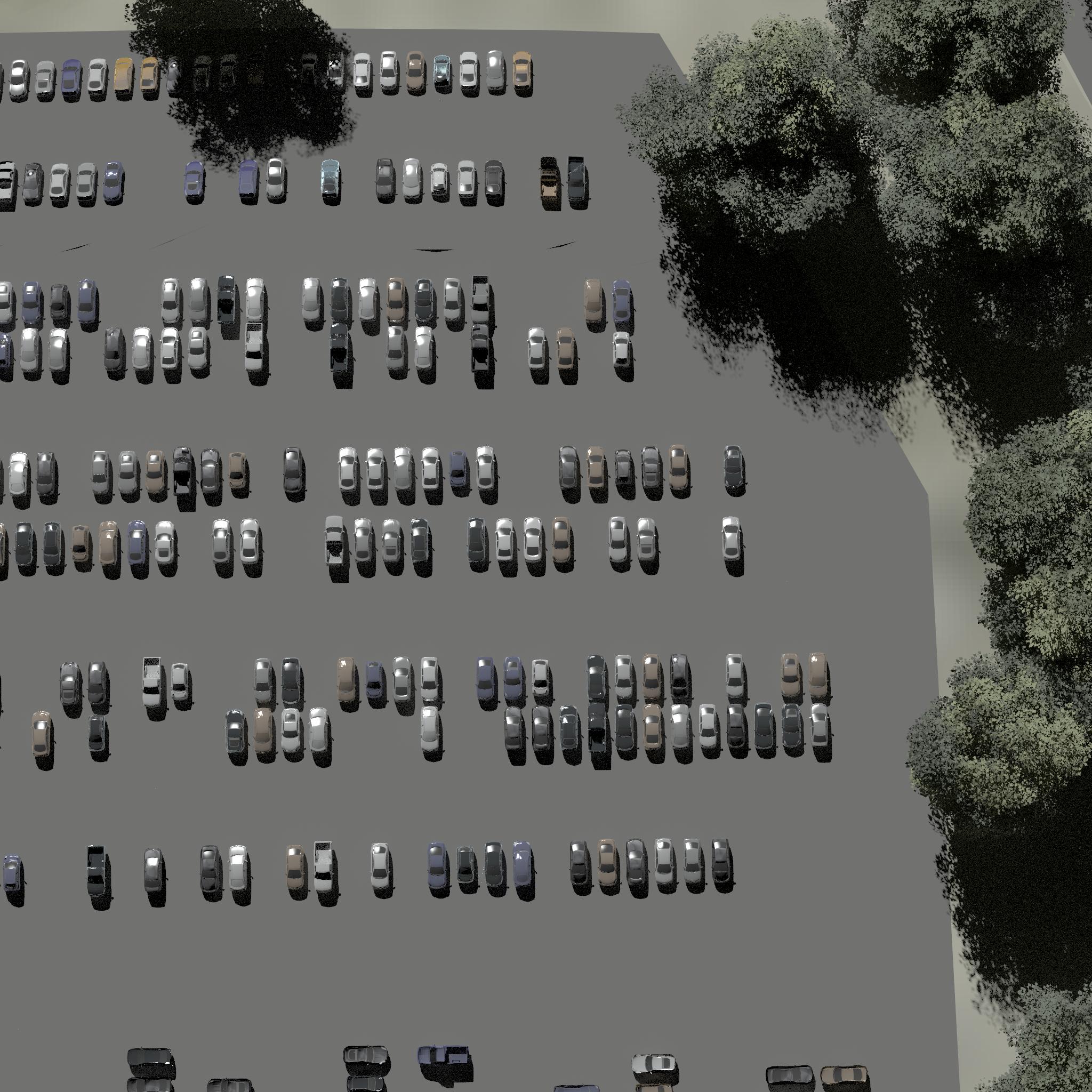}}
    \subcaptionbox{NIR simulation showing fine level of scene detail \label{lot_ir}}
    {\includegraphics[width=0.23\textwidth]{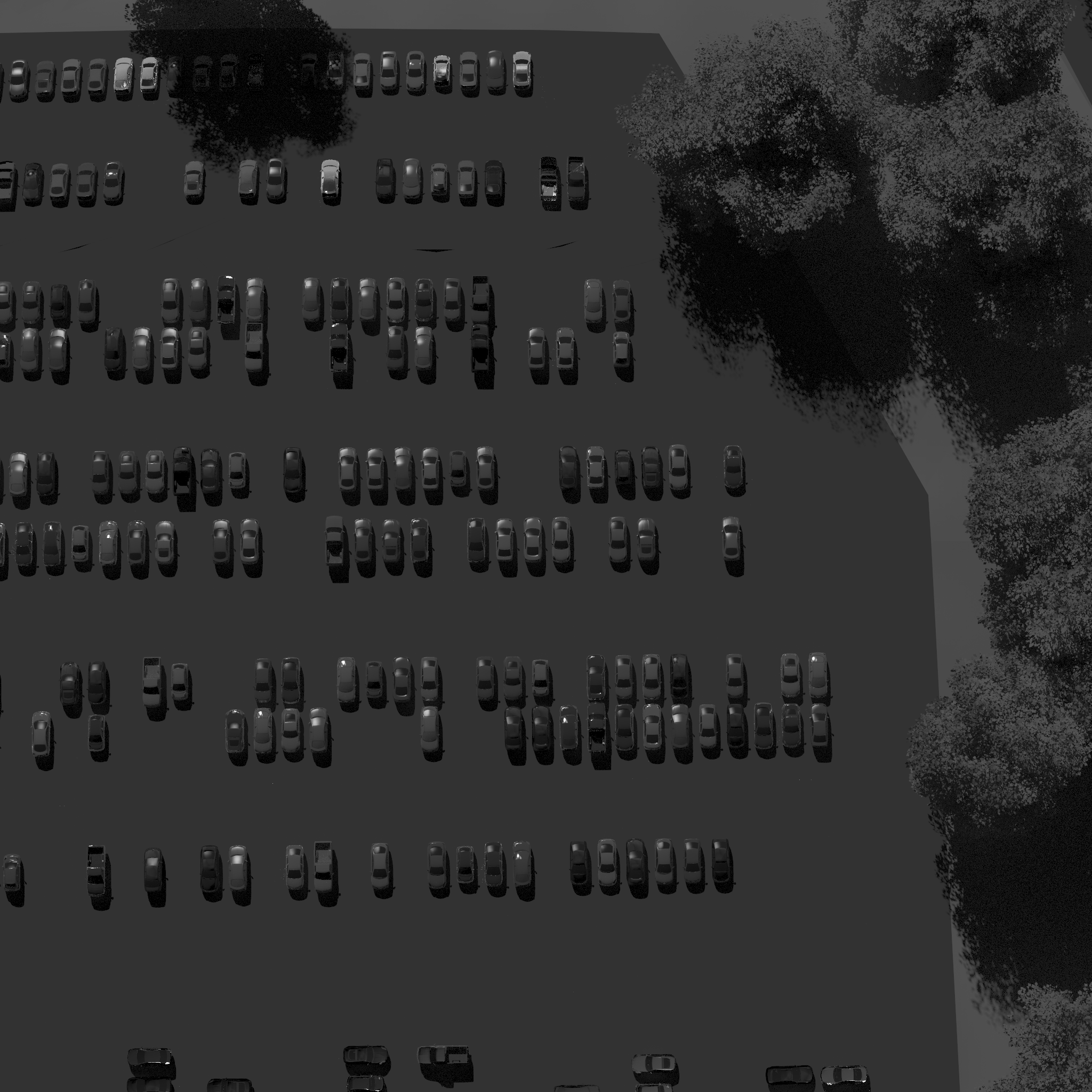}}
    \caption{Simulated imagery at approximately 6x the resolution of the initial satellite imagery.}
    \label{fig:detail_sim}
\end{figure}

One great benefit of generating a simulation in this fashion is the ability to simulate outside of the visible band. We have created a near-infrared simulation of the scene in Fig. \ref{fig:NIR_sim}. Additionally our approach allows us to use geometric elements with high levels of detail as shown in Fig. \ref{fig:detail_sim}, which shows a portion of one of the parking lots in greater detail. In this small scene you can see the trees have individual leaves, the cars are modeled in detail and the vector definition for the parking lot leads to sharp edges at a much higher resolution than the initial material map. The resolution of this detailed scene is roughly six times that of the initial satellite imagery, but benefits greatly from the vector decal maps and populated elements with greater geometric detail. Also notable is the variety of materials on the parked cars. This scene contains 7 unique models of vehicle and 7 different material (paint) variations.

\subsection{Timing Comparison}
The goal of this work has been to create a system to automate the creation of a 3D scene suitable for simulation, and to reduce the amount of manual labor needed to do so. Many parts of the proposed pipeline are difficult to compare to manual processes. An expert scene modeller may be able to quickly edit material maps with image editing software, or create suitably complex 3D models in a CAD or other 3D modelling software. These are skills that are developed over years and difficult to compare to automated approaches. Automation is likely already a part of these workflows, such as the use of procedural modelling. Our approach uses satellite imagery to drive scene creation in a realistic way, allowing novice users to recreate and simulate a specific region of the real world.

One area of our pipeline that can be readily compared is the placement of objects in the scene. Our use of procedural element population uses a set of rules to place vehicles and vegetation in the scene. For the simulated scene in this work we have created rules for how to populate the scene trees, parked cars, and vehicles on the road network. To compare the manual effort needed we have manually populated cars and trees in a small $20m\times20m$ 3D scene. Vehicles were manually placed in a loose parking framework, and trees were positioned around the edge of the lot in a way that looked realistic. Vehicle placement consisted of positioning and rotating so that the cars looked as if they were parked by humans with slight variation in orientation, and some cars backing in and most pulling in forward. Trees were positioned, rotated and scaled to add natural variety to the  appearance.

We allowed 30 minutes for a user familiar with the 3D modelling software to place objects in the scene as realistically as possible. In 30 minutes, 12 cars and 18 trees were added to the scene. This number would likely improve with more experience and practice at scene construction, but is unlikely to change in order of magnitude. Table \ref{table:user_comparison} shows the results of this comparison. Manually populating the scene took roughly one minute per object, whereas through our approach approximately 4670 objects were placed per minute. Our approach completed an area 6875 times larger in less than one third of the time it took to complete the small manual scene. 

\begin{table}[]
\centering
\begin{tabular}{|ll|}
\hline 
\multicolumn{2}{|c|}{\textbf{Manual approach}}\\
Trees placed  & 18                        \\
Cars placed   & 12                        \\
Total Objects & 30                        \\
Time          & 30 minutes                \\
Area covered  & $400m^2$                  \\ \hline \hline
\multicolumn{2}{|c|}{\textbf{Our Approach}} \\
Parked Cars placed    & $3222^*$          \\
Time          & 0.04 seconds              \\ \hline
Cars on Roads placed & 876                \\
Time          & 2.1 seconds               \\ \hline
Trees placed         & 36149              \\
Time          & 8 minutes 35 seconds      \\ \hline
Total objects placed & $40247^*$                     \\
Total time    & 8 minutes 37 seconds      \\
Area Covered  & $2.75km^2$                \\ \hline
\end{tabular}
\caption{A table showing the difference between manual and automated scene population in terms of time spent and quantity of results. \scriptsize$^*$Parked cars denotes spots that can be dynamically populated at runtime by DIRSIG. In practice 70\% of these spots were populated to create a more realistic looking parking lots.}
\label{table:user_comparison}
\end{table}

\section{Conclusion}
In this work we have developed a number of techniques for generating physics driven scene simulation for use with the DIRSIG simulator. Our technique does not make use of lidar, enabling simulations to be constructed for virtually any location on earth, and it greatly decreases manual overhead. Our approach uses commercial satellite imagery for 3D reconstruction, structural modelling, and material classification. We leverage freely available crowd sourced geospatial data to identify roads, parking lots, and foot paths. We use this information as well as material classification to drive the procedural placement of highly detailed objects into our scene. We use a combination of existing spectral libraries as well as custom tools to create rich material definitions for every object and facet of the scene. We create a mixture map for terrain materials that allows for smooth transitions between naturally occurring terrain types without pixelation. Our approach has been developed to allow for realistic simulation at higher levels of detail, and different portions of the spectrum, than is available in the source imagery.


\label{conclusion}

\section{Acknowledgements}
The authors thank the Air Force Research Lab (AFRL) Sensors Directorate for their support of this work through SBIR contract FA8650-18-C-1644. This document is approved for public release via PA Approval \#AFRL-2022-0814. Commercial satellite imagery in the CORE3D public dataset was provided courtesy of Maxar. Dataset was created for the IARPA CORE3D program: https://www.iarpa.gov/index.php/research-programs/core3d \cite{10.1117/12.2304403}.

\bibliographystyle{unsrt}  
\bibliography{references}

\end{document}